\documentclass[runningheads,a4paper]{llncs}

\usepackage{amssymb}
\usepackage{graphicx}

\usepackage{wrapfig}
\usepackage{tabularx}
\usepackage{booktabs}
\usepackage{mathtools}
\mathtoolsset{showonlyrefs,showmanualtags}
\newcommand*{\minimize}{\operatornamewithlimits{minimize}}
\newcommand*{\maximize}{\operatornamewithlimits{maximize}}
\newcommand*{\similarity}{\operatorname{sim}}
\newcommand*{\maxsimc}{\operatorname{maxSimC}}
\newcommand*{\avgsimc}{\operatorname{avgSimC}}

\newcommand*{\avgsim}{\operatorname{avgSim}}

\usepackage{url}
\urldef{\mailsa}\path|{manch043, karypis}@umn.edu|

\begin{document}

\mainmatter  

\title{Distributed representation of multi-sense words: A loss-driven approach}

\titlerunning{Distributed representation of multi-sense words}

%
%
\author{Saurav~Manchanda \and George~Karypis}
%
\authorrunning{Saurav~Manchanda \and George~Karypis}

\institute{University of Minnesota, Twin Cities, MN 55455, USA\\
\mailsa
}

%
%

\toctitle{Lecture Notes in Computer Science}
\tocauthor{Authors' Instructions}
\maketitle

\begin{abstract}
Word2Vec's Skip Gram model is the current state-of-the-art approach for estimating the distributed representation of words. However, it assumes a single vector per word, which is not well-suited for representing words that have multiple senses. This work presents LDMI, a new model for estimating distributional representations of words. LDMI relies on the idea that, if a word carries multiple senses, then having a different representation for each of its senses should lead to a lower loss associated with predicting its co-occurring words, as opposed to the case when a single vector representation is used for all the senses. After identifying the multi-sense words, LDMI clusters the occurrences of these words to assign a sense to each occurrence. Experiments on the contextual word similarity task show that LDMI leads to better performance than competing approaches.
\end{abstract}

\section{Introduction}

Many NLP tasks benefit by embedding the words of a collection into a low dimensional space in a way that captures their syntactic and semantic information. Such NLP tasks include analogy/similarity questions~\cite{mikolov2013efficient}, part-of-speech tagging~\cite{al2013polyglot}, named entity recognition~\cite{al2015polyglot}, machine translation~\cite{zou2013bilingual,mikolov2013exploiting} etc. Distributed representations of words are real-valued, low dimensional embeddings based on the distributional properties of words in large samples of the language data. However, representing each word by a single vector does not properly model the words that have multiple senses (i.e., polysemous and homonymous words). For multi-sense words, a single representation leads to a vector that is the amalgamation of all its different senses, which can lead to ambiguity.

To address this problem, models have been developed to estimate a different representation for each of the senses of multi-sense words. The common idea utilized by these models is that if the words have different senses, then they tend to co-occur with different sets of words. The models proposed by Reisinger and Mooney~\cite{reisinger2010multi}, Huang et al.~\cite{huang2012improving} and the Multiple-Sense Skip-Gram (MSSG) model of Neelakantan et al.~\cite{neelakantan2015efficient} estimates a fixed number of representations per word, without discriminating between the single-sense and multi-sense words. As a result, these approaches fail to identify the right number of senses per word and estimate multiple representations for the words that have a single sense. In addition, these approaches cluster the occurrences without taking into consideration the diversity of words that occur within the contexts of these occurrences (explained in Section \ref{literature}). The Non-Parametric Multiple-Sense Skip-Gram (NP-MSSG) model~\cite{neelakantan2015efficient} estimates a varying number of representations for each word but uses the same clustering approach and hence, is not effective in taking into consideration the diversity of words that occur within the same context.

We present an extension to the Skip-Gram model of Word2Vec to accurately and efficiently estimate a vector representation for each sense of multi-sense words. Our model relies on the fact that, given a word, the Skip-Gram model's loss associated with predicting the words that co-occur with that word, should be greater when that word has multiple senses as compared to the case when it has a single sense. This information is used to identify the words that have multiple senses and estimate a different representation for each of the senses. These representations are estimated using the Skip-Gram model by first clustering the occurrences of the multi-sense words by accounting for the diversity of the words in these contexts. We evaluated the performance of our model for the contextual similarity task on the Stanford's Contextual Word Similarities (SCWS) dataset. When comparing the most likely contextual sense of words, our model was able to achieve approximately 13\% and 10\% improvement over the NP-MSSG and MSSG approaches, respectively. In addition, our qualitative evaluation shows that our model does a better job of identifying the words that have multiple senses over the competing approaches.

\section{Definitions, notations and background} \label{definitions}

Distributed representation of words quantify the syntactic and semantic relations among the words based on their distributional properties in large samples of the language data. The underlying assumption is that the co-occurring words should be similar to each other.  We say that the word $w_j$ co-occurs with the word $w_i$ if $w_j$ occurs within a window around $w_i$. The context of $w_i$ corresponds to the set of words which co-occur with $w_i$ within a window and is represented by $C(w_i)$.

The state-of-the-art technique to learn the distributed representation of words is Word2Vec. The word vector representations produced by Word2Vec are able to capture fine-grained semantic and syntactic regularities in the language data. Word2Vec provides two models to learn word vector representations. The first is the Continuous Bag-of-words Model that involves predicting a word using its context. The second is called the Continuous Skip-gram Model that involves predicting the context using the current word. To estimate the word vectors, Word2Vec trains a simple neural network with a single hidden layer to perform the following task: Given an input word ($w_i$), the network computes the probability for every word in the vocabulary of being in the context of $w_i$. The network is trained such that, if it is given $w_i$ as an input, it will give a higher probability to $w_j$ in the output layer than $w_k$ if $w_j$ occurs in the context of $w_i$ but $w_k$ does not occur in the context of $w_i$. The set of all words in the vocabulary is represented by $V$. The vector associated with the word $w_i$ is denoted by $\boldsymbol{w_i}$. The vector corresponding to word $w_i$ when $w_i$ is used in the context is denoted by $\boldsymbol{\tilde{w_i}}$. The size of the word vector $\boldsymbol{w_i}$ or the context vector $\boldsymbol{\tilde{w_i}}$ is denoted by $d$. 

The objective function for the Skip-Gram model with negative sampling is given by~\cite{goldberg2014word2vec}

\begin{equation}
\minimize\; -\sum\limits_{i=1}^{|V|}\Bigg(\sum\limits_{w_j\in C(w_i)}\log \sigma(\langle\boldsymbol{w_i}, \boldsymbol{\tilde{w_j}}
\rangle)\\+ \sum\limits_{\substack{k \in R(m,|V|)\\ w_k\notin C(w_i)}}\log\sigma(-\langle\boldsymbol{w_i}, \boldsymbol{\tilde{w_k}}\rangle)\Bigg),
\end{equation}
where $R(m,n)$ denotes a set of $m$ random numbers from the range $[1,n]$ (negative samples), $\langle\boldsymbol{w_i, w_j}\rangle$ is the dot product of $\boldsymbol{w_i}$ and $\boldsymbol{w_j}$. and $\sigma(\langle\boldsymbol{w_i}, \boldsymbol{\tilde{w_j}}\rangle)$ is the sigmoid function.

The parameters of the model are estimated using Stochastic Gradient Descent (SGD) in which, for each iteration, the model makes a single pass through every word in the training corpus (say $w_i$) and gathers the context words within a window. The negative samples are sampled from a probability distribution which favors the frequent words.
The model also down-samples the frequent words using a hyper-parameter called the sub-sampling parameter. 

\section{Prior approaches for dealing with multi-sense words}\label{literature}

Various models have been developed to deal with the distributed representations of the multi-sense words. These models presented in this section work by estimating multiple vector-space representations per word, one for each sense. Most of these models estimate a fixed number of vector representations for each word, irrespective of the number of senses associated with a word. In the rest of this section, we review these models and discuss their limitations.

Reisinger and Mooney~\cite{reisinger2010multi} clusters the occurrences of a word using the mixture of von Mises-Fisher distributions~\cite{banerjee2005clustering} clustering method to assign a different sense to each occurrence of the word. The clustering is performed on all the words even if the word has a single sense. This approach estimates a fixed number of vector representations for each word in the vocabulary. As per the authors, the model captures meaningful variation in the word usage and does not assume that each vector representation corresponds to a different sense. Huang et al.~\cite{huang2012improving} also uses the same idea and estimates a fixed number of senses for each word. It uses spherical k-means~\cite{dhillon2001concept} to cluster the occurrences.

Neelakantan et al.~\cite{neelakantan2015efficient} proposed two models built on the top of the Skip-Gram model: Multiple-Sense Skip-Gram (MSSG), and its Non-Parametric counterpart NP-MSSG. MSSG estimates a fixed number of senses per word whereas NP-MSSG discovers varying number of senses. MSSG maintains clusters of the occurrences for each word, each cluster corresponding to a sense. Each occurrence of a word is assigned a sense based on the similarity of its context with the already maintained clusters, and the corresponding vector representation, as well as the sense cluster of the word is updated. During training, NP-MSSG creates a new sense for a word with the probability proportional to the distance of the context to the nearest sense cluster. Both MSSG and NP-MSSG create an auxiliary vector to represent an occurrence, by taking the average of vectors associated with all the words belonging to its context. The similarity between the two occurrences is computed as the cosine similarity between these auxiliary vectors. This approach does not take into consideration the variation among the words that occur within the same context. Another disadvantage is that the auxiliary vector is biased towards the words having higher $L_2$ norm. This leads to noisy clusters, and hence, the senses discovered by these models are not robust.

\section{Loss driven multisense identification (LDMI)}\label{proposed}

In order to address the limitations of the existing models, we developed an extension to the Skip-Gram model that combines two ideas. The first is to identify the multi-sense words and the second is to cluster the occurrences of the identified words such that the clustering correctly accounts for the variation among the words that occur within the same context. We explain these parts as follows:

\subsection{Identifying the words with multiple senses}
For the Skip-Gram model, the loss associated with an occurrence of $w_i$ is
\begin{equation}
L(w_i) = -\Bigg(\sum\limits_{w_j\in C(w_i)}\log \sigma(\langle\boldsymbol{w_i}, \boldsymbol{\tilde{w_j}}
\rangle)\\+ \sum\limits_{\substack{k \in R(m,|V|)\\ w_k\notin C(w_i)}}\log\sigma(-\langle\boldsymbol{w_i}, \boldsymbol{\tilde{w_k}}\rangle)\Bigg).
\end{equation}
The model minimizes $L(w_i)$ by increasing the probability of the co-occurrence of $w_j$ and $w_i$ if $w_j$ is present in the context of $w_i$ and decreasing the probability of the co-occurrence of $w_k$ and $w_i$ if $w_k$ is not present in the context of $w_i$. This happens by aligning the directions of $\boldsymbol{w_i}$ and $\boldsymbol{\tilde{w_j}}$ closer to each other and aligning the directions of $\boldsymbol{w_i}$ and $\boldsymbol{\tilde{w_k}}$ farther from each other. At the end of the optimization process, we expect that the co-occurring words have their vectors aligned closer in the vector space. However, consider the polysemous word \textit{bat}. We expect that the vector representation of \textit{bat} is aligned in a direction closer to the directions of the vectors representing the terms like \textit{ball, baseball,  sports} etc. (the sense corresponding to \textit{sports}). We also expect that the vector representation of \textit{bat} is aligned in a direction closer to the directions of the vectors representing the terms like \textit{animal, batman, nocturnal} etc. (the sense corresponding to \textit{animals}). But at the same time, we do not expect that the directions of the vectors representing the words corresponding to the \textit{sports} sense are closer to the directions of the vectors representing the words corresponding to the \textit{animal} sense. This leads to the direction of the vector representing \textit{bat} lying in between the directions of the vectors representing the words corresponding to the \textit{sports} sense and the directions of the vectors representing the words corresponding to the \textit{animal} sense. Consequently, the multi-sense words will tend to contribute more to the overall loss than the words with a single sense.

Having a vector representation for each sense of the word \textit{bat} will avoid this scenario, as each sense can be considered as a new single-sense word in the vocabulary. Hence, the loss associated with a word provides us information regarding whether a word has multiple senses or not. LDMI leverages this insight to identify a word $w_i$ as multi-sense if the average $L(w_i)$ across all its occurrences is more than a threshold. However, $L(w_i)$ has a random component associated with it, in the form of negative samples. We found that, in general, infrequent words have higher loss as compared to the frequent words. This can be attributed to the fact that given a random negative sample while calculating the loss, there is a greater chance that the frequent words have already seen this negative sample before during the optimization process as compared to the infrequent words. This way, infrequent words end up having higher loss than frequent words. Therefore, for the selection purposes, we ignore the loss associated with negative samples. We denote the average loss associated with the prediction of the context words in an occurrence of $w_i$ as $L^+(w_i)$ and define it as

\begin{equation}
L^+(w_i) = -\frac{1}{|C(w_i)|}\sum\limits_{w_j\in C(w_i)}\log \sigma(\langle\boldsymbol{w_i}, \boldsymbol{\tilde{w_j}}\rangle).
\end{equation}
We describe $L^+(w_i)$ as the \textit{contextual loss} associated with an occurrence of $w_i$. 

To identify the multi-sense words, LDMI performs a few iterations to optimize the loss function on the text dataset, and shortlist the words with average contextual loss (average $L^+(w_i)$ across all the occurrences of the $w_i$) that is higher than a threshold. These shortlisted words represent the identified multi-sense words, which form the input of the second step described in the next section.

\subsection{Clustering the occurrences}

To assign senses to the occurrences of each of the identified multi-sense words, LDMI clusters its occurrences so that each of the clusters corresponds to a particular sense. The clustering solution employs the $\mathcal{I}_1$ criterion function~\cite{karypis2002cluto} which maximizes the objective function of the form

\begin{equation}\label{eq:objective}
\maximize \sum\limits_{i = 1}^{k}n_iQ(S_i),
\end{equation}
where $Q(S_i)$ is the quality of cluster $S_i$ whose size is $n_i$. We define $Q(S_i)$ as

\begin{equation}\label{eq:quality}
Q(S_i) = \frac{1}{n_i^2}\sum\limits_{u,v \in S_i}\similarity(u, v),
\end{equation}
where $\similarity(u, v)$ denotes the similarity between the occurrences $u$ and $v$, and is given by

\begin{equation}\label{eq:context_sim}
\similarity(u, v) = \frac{1}{|C(u)||C(v)|}\sum\limits_{x \in C(u)}\sum\limits_{y \in C(v)}\cos(x, y),
\end{equation}
According to Equation \eqref{eq:context_sim}, LDMI measures the similarity between the two occurrences as the average of the pairwise cosine similarities between the words belonging to the contexts of these occurrences. This approach considers the variation among the words that occur within the same context. We can simplify Equation \eqref{eq:objective} to the following equation

\begin{equation}
\maximize \sum\limits_{i = 1}^{k}\frac{1}{n_i} \left\lVert\sum\limits_{u \in S_i}\left(\sum\limits_{x \in C(u)}\frac{\boldsymbol{x}}{\lVert\boldsymbol{x}\rVert_2}\right)\right\rVert_2^2.
\end{equation}
LDMI maximizes this objective function using a greedy incremental strategy~\cite{karypis2002cluto}.

\subsection{Putting everything together}

LDMI is an iterative algorithm with two steps in each iteration. The first step is to perform a few SGD iterations to optimize the loss function. In the second step, it calculates the contextual loss associated with each occurrence of each word and identifies the words having the average contextual loss that is more than a threshold. It then clusters the occurrences of the identified multi-sense words into two clusters ($k = 2$) as per the clustering approach discussed earlier. The algorithm terminates after a fixed number of iterations. $x$ number of iterations of LDMI can estimate a maximum of $2^x$ senses for each word. 

\section{Experimental methodology}\label{experiments}

\subsection{Datasets}

We train LDMI on two corpora of varying sizes: The Wall Street Journal (WSJ) dataset~\cite{harman1993tipster} and the Google's One Billion Word (GOBW) dataset~\cite{chelba2013one}. In preprocessing, we removed all the words which contained a number or did not contain any alphabet and converted the remaining words to lower case.

\begin{wraptable}{r}{0.48\columnwidth}
\small
\centering
  \caption{Dataset statistics.}
  \begin{tabularx}{2.3in}{Xrr}
    \hline
    Dataset&Vocabulary size&Total words\\
    \hline
    WSJ&88,118&62,653,821\\
    GOBW&73,443&710,848,599\\
    \hline
\end{tabularx}
  \label{tab:dataset}
\end{wraptable}For WSJ, we removed all the words with frequency less than 10 and for GOBW, we removed all the words with frequency less than 100. The statistics of these datasets after preprocessing are presented in Table \ref{tab:dataset}. 

We use Stanford's Contextual Word Similarities (SCWS) dataset~\cite{huang2012improving} for evaluation on the contextual word similarity task. SCWS contains human judgments on pairs of words (2,003 in total) presented in sentential context. The word pairs are chosen so as to reflect interesting variations in meanings.

When the contextual information is not present, different people can consider different senses when giving a similarity judgment. Therefore, having representations for all the senses of a word can help us to find similarities which align better with the human judgments, as compared to having a single representation of a word. To investigate this, we evaluated our model on the WordSim-353 dataset~\cite{finkelstein2001placing}, which consists of 353 pairs of nouns, without any contextual information. Each pair has an associated averaged human judgments on similarity.

\subsection{Evaluation methodology and metrics} 

\subsubsection{Baselines.}
We compare the LDMI model with the MSSG and NP-MSSG approaches as they are also built on top of the Skip-Gram model. As mentioned earlier, MSSG estimates the vectors for a fixed number of senses per word whereas NP-MSSG discovers varying number of senses per word. To illustrate the advantage of using the clustering with $\mathcal{I}_1$ criterion over the clustering approach used by the competing models, we also compare LDMI with LDMI-SK. LDMI-SK uses the same approach to select the multi-sense words as used by the LDMI, but instead of clustering with the $\mathcal{I}_1$ criterion, it uses spherical K-means~\cite{dhillon2001concept}.

\subsubsection{Parameter selection.}

\begin{wrapfigure}{r}{0.45\columnwidth}
\belowcaptionskip = -22pt
\abovecaptionskip = -2pt
\small
\centering
\includegraphics[width=2.2in]{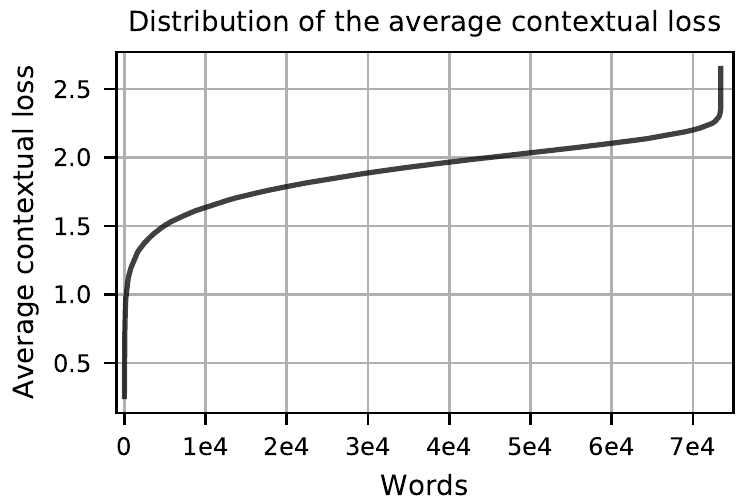}
\caption{Distribution of the average contextual loss for all words (Words on the x-axis are sorted in order of their loss)}
\label{fig:contextual_loss}
\end{wrapfigure}

For all our experiments, we consider 10 negative samples and a symmetric window of 10 words. The sub-sampling parameter is $10^{-4}$ for both the datasets. To avoid clustering the infrequent and stop-words, we only consider the words within a frequency range to select them as the multi-sense words. For the WSJ dataset, we consider the words with frequency between 50 and 30,000 while for the GOBW dataset, we consider the words with frequency between 500 and 300,000. For the WSJ dataset, we consider only 50-dimensional embeddings while for the GOBW dataset, we consider $50$, $100$ and $200$ dimensional embeddings. The model checks for multi-sense words after every 5 iterations. We selected our hyperparameter values by a small amount of manual exploration to get the best performing model. To decide the threshold for the average contextual loss to select the multi-sense words, we consider the distribution of the average contextual loss after running an iteration of Skip-Gram. For example, Fig. \ref{fig:contextual_loss} shows the average contextual loss of every word in the vocabulary for the GOBW dataset for the $50$-dimensional embeddings. We can see that there is an increase in the average contextual loss around $2.0-2.4$.
 We experiment around this range to select a loss threshold for which our model performs best. For the experiments presented in this paper, this threshold is set to $2.15$ for the WSJ (50-dimensional embeddings), and $2.15$, $2.10$ and $2.05$ for the GOBW corresponding to the 50, 100 and 200-dimensional embeddings, respectively. With increasing dimensionality of the vectors, we are able to model the information from the dataset in a better way, which leads to a relatively lower loss.

For the MSSG and NP-MSSG models, we use the same hyperparameter values as used by Neelakantan et al.~\cite{neelakantan2015efficient}. For MSSG, the number of senses is set to 3. Increasing the number of senses involves a compromise between getting the correct number of senses for some words while noisy senses for the others. For NP-MSSG, the maximum number of senses is set to $10$ and the parameter $\lambda$ is set to $-0.5$ (A new sense cluster is created if the similarity of an occurrence to the existing sense clusters is less than $\lambda$). The models are trained using SGD with AdaGrad~\cite{duchi2011adaptive} with $0.025$ as the initial learning rate and we run 15 iterations. 

\subsubsection{Metrics.}
For evaluation, we use the similarities calculated by our model and sort them to create an ordering among all the word-pairs. We compare this ordering against the one obtained by the human judgments. To do this comparison, we use the Spearman rank correlation ($\rho$). Higher score for the Spearman rank correlation corresponds to the better correlation between the respective orderings. For the SCWS dataset, to measure the similarity between two words given their sentential contexts, we use two different metrics~\cite{reisinger2010multi}. 
The first is the $\maxsimc$, which for each word in the pair, identifies the sense of the word that is the most similar to its context and then compares those two senses. It is computed as
\begin{equation}
\maxsimc(w_1, w_2, C(w_1), C(w_2)) \\= \cos(\hat{\pi}(w_1), \hat{\pi}(w_2)),
\end{equation}
where, $\hat{\pi}(w_i)$ is the vector representation of the sense that is most similar to $C(w_i)$. As in Equation \eqref{eq:context_sim}, we measure the similarity between $x$ and $C(w_i)$ as

 \begin{equation}
\similarity(x, C(w_i)) = \frac{1}{Z}\left(\sum\limits_{y\in C(w_i)}\sum\limits_{j=1}^{m(y)}\cos(x,V(y, j))\right),
\end{equation}
where, $Z = \sum_{y\in C(w_i)}m(y)$, $m(y)$ is the number of senses discovered for the word $y$ and $V(y,i)$ is the vector representation associated with the $i$th sense of the word $y$. For simplicity, we consider all the senses of the words in the sentential context for the similarity calculation. The second metric is the $\avgsimc$ which calculates the similarity between the two words as the weighed average of the similarities between each of their senses. It is computed as

\begin{multline}
\avgsimc(w_1, w_2, C(w_1), C(w_2)) = \\\sum\limits_{i=1}^{m(w_1)}\sum\limits_{j=1}^{m(w_2)}\Bigg(Pr(w_1, i, C(w_1))Pr(w_2, j, C(w_2)) 
\times\cos(V(w_1, i), V(w_2, j))\Bigg),
\end{multline}
where $Pr(x,i,C(x))$ is the probability that $x$ takes the $i$th sense given the context $C(x)$. We calculate $Pr(x,i,C(x))$ as

\begin{equation}
Pr(x,i,C(x)) = \frac{1}{N}\left(\frac{1}{1-\similarity(x,C(x))}\right),
\end{equation}
where $N$ is the normalization constant so that the probabilities add to $1$. Note that, the $\maxsimc$ metric models the similarity between two words with respect to the most probable identified sense for each of them. If there are noisy senses as a result of overclustering, $\maxsimc$ will penalize them. Hence, $\maxsimc$ is a stricter metric as compared to the $\avgsimc$.

For the WordSim-353 dataset, we used the $\avgsim$ metric, which is qualitatively similar to the $\avgsimc$, but does not take contextual information into consideration. The $\avgsim$ metric is calculated as

\begin{equation}
\avgsim(w_1, w_2) = \frac{1}{m(w_1)m(w_2)}\\\times\sum\limits_{i=1}^{m(w_1)}\sum\limits_{j=1}^{m(w_2)}\cos(V(w_1, i), V(w_2, j)).
\end{equation}
For qualitative analysis, we look into the similar words corresponding to different senses for some of the words identified as multi-sense by the LDMI and compare them to the ones discovered by the competing approaches. 

\section{Results and discussion}\label{results}

\subsection{Quantitative analysis}
\begin{table*}[!t]
\small
\centering
  \caption{Results for the Spearman rank correlation ($\rho \times 100$).}
  \begin{tabular*}{\textwidth}{l@{\extracolsep{\fill}}llrrrr}
    \hline
    Dataset&Model&$d$&$\maxsimc$&$\avgsimc$&$\avgsim$\\
    &&&(SCWS)&(SCWS)&(WordSim-353)\\
    \hline
    WSJ&Skip-Gram&50&57.0&57.0&54.9\\
    WSJ&MSSG&50&41.4&56.3&50.5\\
    WSJ&NP-MSSG&50&33.0&52.2&47.4\\
    WSJ&LDMI-SK&50&57.1&57.9&55.2\\
    WSJ&LDMI&50&\textbf{57.9}&\textbf{58.9}&\textbf{56.8}\\
    \hline
    GOBW&Skip-Gram&50&60.1&60.1&62.0\\
    GOBW&MSSG&50&50.0&59.6&57.1\\
    GOBW&NP-MSSG&50&48.2&60.0&58.9\\
    GOBW&LDMI-SK&50&60.1&60.6&62.8\\
    GOBW&LDMI&50&\textbf{60.6}&\textbf{61.2}&\textbf{63.8}\\
    \hline
    GOBW&Skip-Gram&100&61.7&61.7&64.3\\
    GOBW&MSSG&100&53.4&62.6&60.4\\
    GOBW&NP-MSSG&100&47.9&\textbf{63.3}&61.7\\
    GOBW&LDMI-SK&100&61.9&62.4&64.9\\
    GOBW&LDMI&100&\textbf{62.2}&63.1&\textbf{65.3}\\
    \hline
    GOBW&Skip-Gram&200&63.1&63.1&65.4\\
    GOBW&MSSG&200&54.7&64.0&64.2\\
    GOBW&NP-MSSG&200&51.5&64.1&62.8\\
    GOBW&LDMI-SK&200&63.3&63.9&66.4\\
    GOBW&LDMI&200&\textbf{63.9}&\textbf{64.4}&\textbf{66.8}\\
    \hline
\end{tabular*}
  \label{tab:scws_spearman}
\end{table*}
Table \ref{tab:scws_spearman} shows the Spearman rank correlation ($\rho$) on the SCWS and WordSim-353 dataset for various models and different vector dimensions. For all the vector dimensions, LDMI performs better than the competing approaches on the $\maxsimc$ metric. For the GOBW dataset, LDMI shows an average improvement of about 13\% over the NP-MSSG and 10\% over the MSSG on the $\maxsimc$ metric. The average is taken over all vector dimensions. This shows the advantage of LDMI over the competing approaches. For the $\avgsimc$ metric, LDMI performs at par with the competing approaches. The other approaches are not as effective in identifying the correct number of senses, leading to noisy clusters and hence, poor performance on the $\maxsimc$ metric. LDMI also performs better than LDMI-SK on both $\maxsimc$ and $\avgsimc$, demonstrating the effectiveness of the clustering approach employed by LDMI over spherical k-means. 

\begin{wrapfigure}{r}{0.50\columnwidth}
\belowcaptionskip = -20pt
\abovecaptionskip = -2pt
\small
\centering
\includegraphics[width=2.5in]{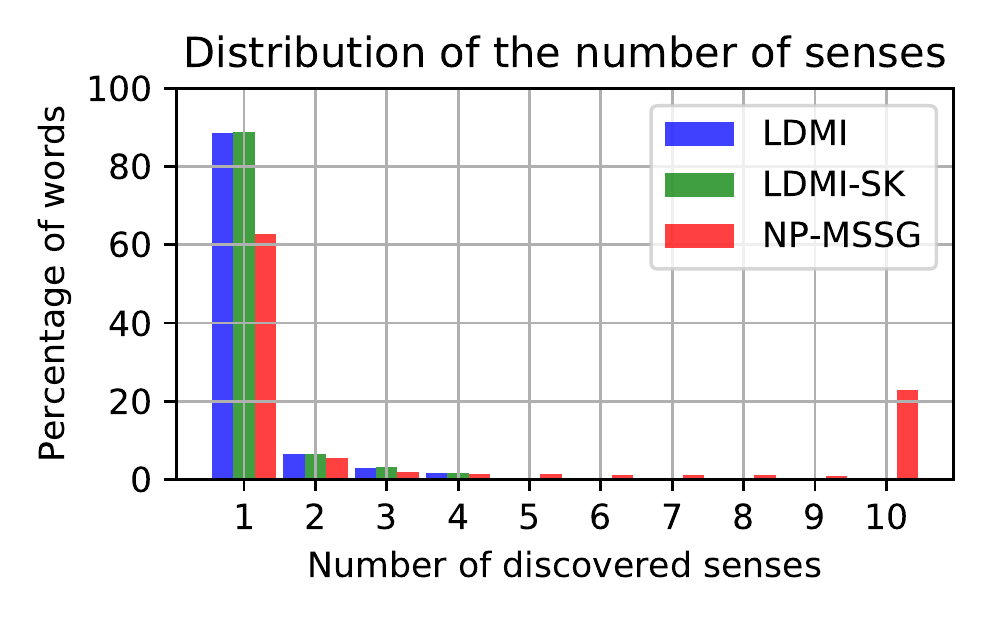}
\caption{Distribution of the number of senses}
\label{fig:sense_distributon}
\end{wrapfigure} 
Similarly, LDMI performs better than other approaches on the $\avgsim$ metric for the WordSim-353 dataset in all the cases, further demonstrating the advantage of LDMI. 
Fig. \ref{fig:sense_distributon} shows the distribution of the number of senses discovered by the LDMI, LDMI-SK and NP-MSSG model for the GOBW dataset and 200-dimensional embeddings. We can see that LDMI and LDMI-SK discover $88\%$ of the words as single-sense, while NP-MSSG discovers $63\%$ of the words as single-sense.

In addition, we used the Kolmogorov-Smirnov two-sample test to assess if LDMI's performance advantage over the Skip-Gram is statistically significant. We performed the test on $\maxsimc$ and $\avgsimc$ metrics corresponding to the 1,000 runs each of LDMI and Skip-Gram on the WSJ dataset. For the null hypothesis that the two samples are derived from the same distribution, the resulting p-value ($\approx 10^{-8}$) shows that the difference is statistically significant for both $\maxsimc$ and $\avgsimc$ metrics. Similarly, the difference in the LDMI's and LDMI-SK's performance is also found to be statistically significant.

\subsection{Qualitative analysis}
In order to evaluate the actual senses that the different models identify, we look into the similar words corresponding to different senses for some of the words identified as multi-sense by LDMI. We compare these discovered senses with other competing approaches. Table \ref{tab:similar_words} shows the similar words (corresponding to the cosine similarity) with respect to some of the words that LDMI identified as multi-sense words and estimated a different vector representation for each sense. The results correspond to the 50-dimensional embeddings for the GOBW dataset. The table illustrates that LDMI is able to identify meaningful senses. For example, it is able to identify two senses of the word \textit{digest}, one corresponding to the \textit{food} sense and the other to the \textit{magazine} sense. For the word \textit{block}, it is able to identify two senses, corresponding to the \textit{hindrance} and \textit{address} sense.

\begin{table*}[!t]
\small
\centering 
\caption{Top similar words for different senses of the multi-sense words (different lines in a row correspond to different senses).}
  \begin{tabularx}{\linewidth}{lXl}
    \hline
    Word&Similar words&Sense\\
    \hline
    figure& status; considered; iconoclast; charismatic; stature; known; &leader\\
    & calculate; understand; know; find; quantify; explain; how; tell; &deduce\\
    & doubling; tenth; average; percentage; total; cent; gdp; estimate&numbers\\
    \hline
    cool& breezy; gentle; chill; hot; warm; chilled; cooler; sunny; frosty; &weather\\
    & pretty; liking; classy; quite; nice; wise; fast; nicer; okay; mad;&expression\\
  \hline
    block& amend; revoke; disallow; overturn; thwart; nullify; reject; &hindrance\\
    & alley; avenue; waterside; duplex; opposite; lane; boulevard;&address\\
  \hline
    digest& eat; metabolize; starches; reproduce; chew; gut; consume; &food\\
    & editor; guide; penguin; publisher; compilers; editions; paper;&magazine\\
  \hline
    head& arm; shoulder; ankles; neck; throat; torso; nose; limp; toe; &body\\
    & assistant; associate; deputy; chief; vice; executive; adviser; &organization\\
  \hline
\end{tabularx}
\label{tab:similar_words}
\end{table*}
Table \ref{tab:similar_words_comparison} shows the similar words with respect to the identified senses for the words \textit{digest} and \textit{block} by the competing approaches. We can see that LDMI is able to identify more comprehensible senses for \textit{digest} and \textit{block}, compared to MSSG and NP-MSSG. Compared to the  LDMI, LDMI-SK finds redundant senses for the word \textit{digest}, but overall, the senses found by the LDMI-SK are comparable to the ones found by the LDMI. 

\begin{table}[!t]
\small
\centering  
  \caption{Senses discovered by the competing approaches (different lines in a row correspond to different senses).}
  \begin{tabularx}{\columnwidth}{X|X}
    \hline
    \textbf{digest (Skip-Gram)}&\textbf{block (Skip-Gram)}\\
    nutritional; publishes; bittman; reader&annex; barricade; snaked; curving; narrow\\
    \hline
    \textbf{digest (MSSG)}&\textbf{block (MSSG)}\\
    comenu; ponder; catch; turn; ignore&street; corner; brick; lofts; lombard; wall\\
	areat; grow; tease; releasing; warts&yancey; linden; calif; stapleton; spruce; ellis\\
	nast; conde; blender; magazine; edition&bypass; allow; clears; compel; stop\\
    \hline
    \textbf{digest (NP-MSSG)}&\textbf{block (NP-MSSG)}\\
    guide; bible; ebook; danielle; bookseller&acquire; pipeline; blocks; stumbling; owner\\
	snippets; find; squeeze; analyze; tease&override; approve; thwart; strip; overturn\\
	eat; ingest; starches; microbes; produce&townhouse; alley; blocks; street; entrance\\
	oprah; cosmopolitan; editor; conde; nast&mill; dix; pickens; dewitt; woodland; lane\\
	disappointing; ahead; unease; nervousness&slices; rebounded; wrestled; effort; limit\\
	observer; writing; irina; reveals; bewildered&target; remove; hamper; remove; binding\\
	&hinder; reclaim; thwart; hamper; stop\\
	&side; blocks; stand; walls; concrete; front\\
	&approve; enforce; overturn; halted; delay\\
	&inside; simply; retrieve; track; stopping\\
    \hline
    \textbf{digest (LDMI-SK)}&\textbf{block (LDMI-SK)}\\
    almanac; deloitte; nast; wired; guide&cinder; fronted; avenue; flagstone;bricks\\
	sugars; bacteria; ingest; enzymes; nutrients&amend; blocking; withhold; bypass; stall\\
	liking; sort; swallow; find; bite; whole&\\
	find; fresh; percolate; tease; answers&\\
    \hline
	\end{tabularx}
  \label{tab:similar_words_comparison}
\end{table}

\section{Conclusion}\label{conclusion}
We presented LDMI, a model to estimate distributed representations of the multi-sense words. LDMI is able to efficiently identify the meaningful senses of words and estimate the vector embeddings for each sense of these identified words. The vector embeddings produced by LDMI achieves state-of-the-art results on the contextual similarity task by outperforming the other related work. 

\subsubsection{Acknowledgments.}
This work was supported in part by NSF (IIS-1247632, IIP-1414153, IIS-1447788, IIS-1704074, CNS-1757916), Army Research Office (W911NF-14-1-0316), Intel Software and Services Group, and the Digital Technology Center at the University of Minnesota. Access to research and computing facilities was provided by the Digital Technology Center and the Minnesota Supercomputing Institute. This is a pre-print of an article published in PAKDD 2018: Advances in Knowledge Discovery and Data Mining, as a part of the Lecture Notes in Computer Science book series (LNCS, volume 10938). The final authenticated version is available online at: https://doi.org/10.1007/978-3-319-93037-4\_27.

%
%
\bibliographystyle{splncs03}
\bibliography{refs}

\begin{thebibliography}{10}
\providecommand{\url}[1]{\texttt{#1}}
\providecommand{\urlprefix}{URL }

\bibitem{al2015polyglot}
Al-Rfou, R., Kulkarni, V., Perozzi, B., Skiena, S.: Polyglot-ner: Massive
  multilingual named entity recognition. In: SDM (2015)

\bibitem{al2013polyglot}
Al-Rfou, R., Perozzi, B., Skiena, S.: Polyglot: Distributed word
  representations for multilingual nlp. CoNLL  (2013)

\bibitem{banerjee2005clustering}
Banerjee, A., Dhillon, I.S., Ghosh, J., Sra, S.: Clustering on the unit
  hypersphere using von mises-fisher distributions. Journal of Machine Learning
  Research  6(Sep) (2005)

\bibitem{chelba2013one}
Chelba, C., Mikolov, T., Schuster, M., Ge, Q., Brants, T., Koehn, P., Robinson,
  T.: One billion word benchmark for measuring progress in statistical language
  modeling. Tech. rep. (2013), \url{http://arxiv.org/abs/1312.3005}

\bibitem{dhillon2001concept}
Dhillon, I.S., Modha, D.S.: Concept decompositions for large sparse text data
  using clustering. Machine learning  42(1-2) (2001)

\bibitem{duchi2011adaptive}
Duchi, J., Hazan, E., Singer, Y.: Adaptive subgradient methods for online
  learning and stochastic optimization. Journal of Machine Learning Research
  12(Jul) (2011)

\bibitem{finkelstein2001placing}
Finkelstein, L., Gabrilovich, E., Matias, Y., Rivlin, E., Solan, Z., Wolfman,
  G., Ruppin, E.: Placing search in context: The concept revisited. In:
  Proceedings of the 10th international conference on World Wide Web (2001)

\bibitem{goldberg2014word2vec}
Goldberg, Y., Levy, O.: word2vec explained: Deriving mikolov et al.'s
  negative-sampling word-embedding method. arXiv preprint arXiv:1402.3722
  (2014)

\bibitem{harman1993tipster}
Harman, D., Liberman, M.: Tipster complete. Corpus number LDC93T3A, Linguistic
  Data Consortium, Philadelphia  (1993)

\bibitem{huang2012improving}
Huang, E.H., Socher, R., Manning, C.D., Ng, A.Y.: Improving word
  representations via global context and multiple word prototypes. In:
  Proceedings of the 50th Annual Meeting of the Association for Computational
  Linguistics (2012)

\bibitem{mikolov2013efficient}
Mikolov, T., Chen, K., Corrado, G., Dean, J.: Efficient estimation of word
  representations in vector space. arXiv preprint arXiv:1301.3781  (2013)

\bibitem{mikolov2013exploiting}
Mikolov, T., Le, Q.V., Sutskever, I.: Exploiting similarities among languages
  for machine translation. arXiv preprint arXiv:1309.4168  (2013)

\bibitem{neelakantan2015efficient}
Neelakantan, A., Shankar, J., Passos, A., McCallum, A.: Efficient
  non-parametric estimation of multiple embeddings per word in vector space.
  In: EMNLP (2014)

\bibitem{reisinger2010multi}
Reisinger, J., Mooney, R.J.: Multi-prototype vector-space models of word
  meaning. In: NAACL:HLT (2010)

\bibitem{karypis2002cluto}
Zhao, Y., Karypis, G.: Criterion functions for document clustering. Tech. rep.,
  Department of Computer Science, University of Minnesota (2005)

\bibitem{zou2013bilingual}
Zou, W.Y., Socher, R., Cer, D.M., Manning, C.D.: Bilingual word embeddings for
  phrase-based machine translation. In: EMNLP (2013)

\end{thebibliography}
\end{document}